\documentclass[letterpaper]{article} 
\usepackage{aaai25}  
\usepackage{times}  
\usepackage{helvet}  
\usepackage{courier}  
\usepackage[hyphens]{url}  
\usepackage{graphicx} 
\usepackage{natbib}  
\usepackage{caption} 
\usepackage{algorithm}
\usepackage{algorithmic}
\usepackage{amsmath}
\usepackage{amssymb}
\usepackage{float}
\usepackage{newfloat}
\usepackage{listings}
\DeclareCaptionStyle{ruled}{labelfont=normalfont,labelsep=colon,strut=off} 
\floatstyle{ruled}
\newfloat{listing}{tb}{lst}{}
\floatname{listing}{Listing}
\title{Single Image Rolling Shutter Removal with Diffusion Models}
\author{
	Zhanglei Yang\textsuperscript{\rm 1,\rm 2}\equalcontrib, Haipeng Li\textsuperscript{\rm 1,\rm 2}\equalcontrib, Mingbo Hong\textsuperscript{\rm 2}, Chen-Lin Zhang\textsuperscript{\rm 3}, Jiajun Li\textsuperscript{\rm 4}, Shuaicheng Liu\textsuperscript{\rm 1,\rm 2,\footnote{Corresponding author}}
}
\affiliations{
	\textsuperscript{\rm 1}University of Electronic Science and Technology of China\\
	\textsuperscript{\rm 2}Megvii Technology\\
	\textsuperscript{\rm 3} Moonshot AI \\
        \textsuperscript{\rm 4} Noumena AI \\
        yangzl11974@gmail.com,
	\{lihaipeng@std.,liushuaicheng@\}uestc.edu.cn\\mingbohong97@gmail.com, zclnjucs@gmail.com, 
         taringlee@noumena.com.cn
        
}

\usepackage{bibentry}

\begin{document}

\maketitle

\begin{abstract}
We present RS-Diffusion, the first Diffusion Models-based method for single-frame Rolling Shutter (RS) correction. RS artifacts compromise visual quality of frames due to the row-wise exposure of CMOS sensors. Most previous methods have focused on multi-frame approaches, using temporal information from consecutive frames for the motion rectification. However, few approaches address the more challenging but important single frame RS correction. In this work, we present an ``image-to-motion" framework via diffusion techniques, with a designed patch-attention module. In addition, we present the RS-Real dataset, comprised of captured RS frames alongside their corresponding Global Shutter (GS) ground-truth pairs. The GS frames are corrected from the RS ones, guided by the corresponding Inertial Measurement Unit (IMU) gyroscope data acquired during capture. Experiments show that RS-Diffusion surpasses previous single-frame RS methods, demonstrates the potential of diffusion-based approaches, and provides a valuable dataset for further research.

\end{abstract}


\begin{links}
\link{Code}{https://github.com/lhaippp/RS-Diffusion}
\link{Datasets}{https://huggingface.co/Lhaippp/RS-Diffusion}
\end{links}

\section{Introduction}
\label{sec:intro}

\begin{figure}[t]
\begin{center}
     \includegraphics[width=1\linewidth]{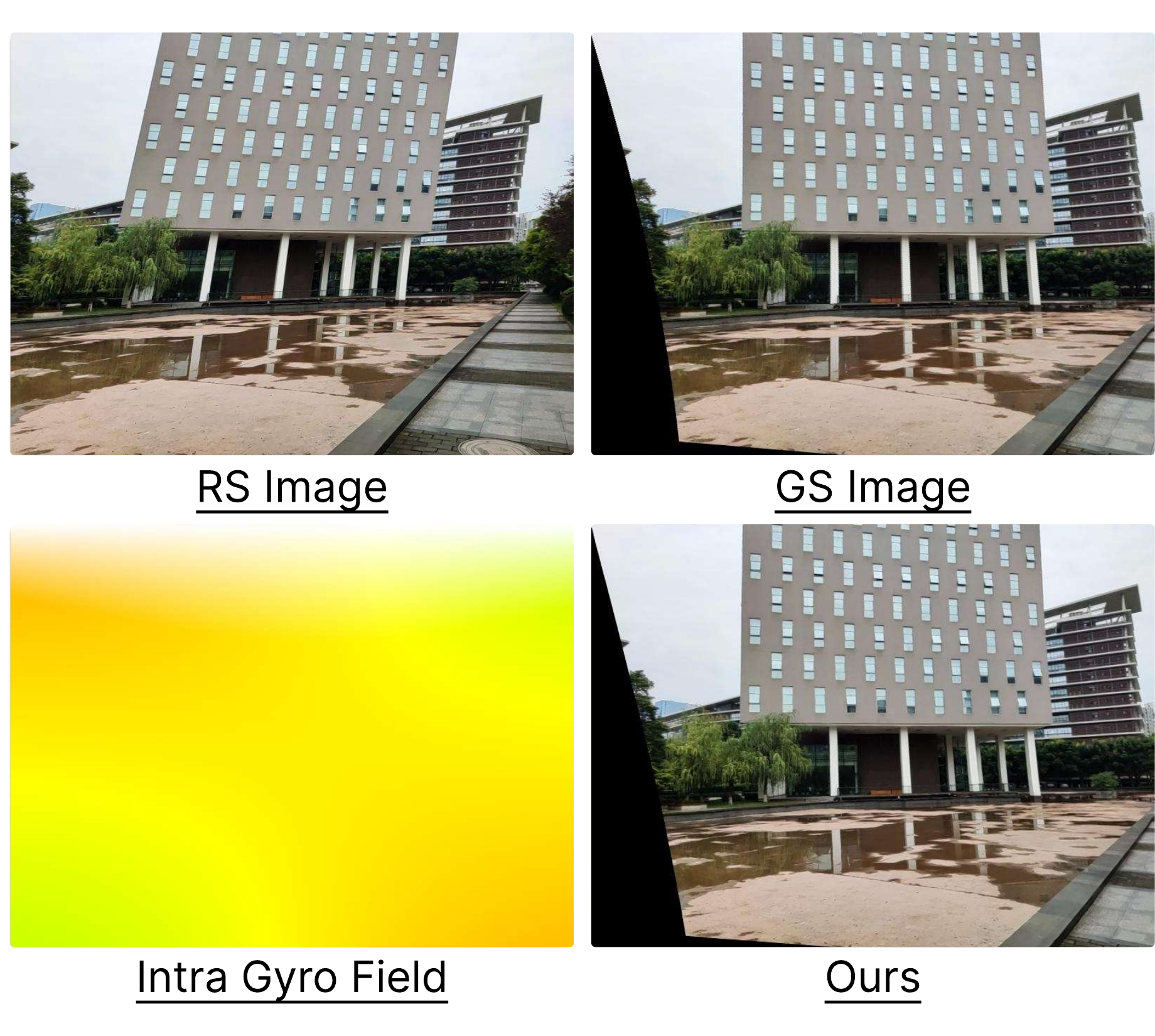}
\end{center}
\caption{Illustration of the proposed dataset and our results. The first row features a rolling-shutter (RS) image captured in realistic scenes, along with the corresponding ground-truth global-shutter (GS) image. The ground-truth flow used for correcting the RS image to the GS image is displayed on the left side of the second row. In the bottom right, we showcase our corrected RS image.}
\label{fig:teaser}
\end{figure}

Rolling shutter (RS) is a common effect encountered when capturing images with CMOS sensors. It results from varying exposure times for different rows in each frame, causing artifacts like distorted straight lines and skewed image content, as shown in Fig.~\ref{fig:teaser} (RS Image). These distortions are not only visually unpleasing but also detrimental to downstream tasks~\cite{hedborg2012rolling,saurer2016sparse,albl2015r6p,saurer2013rolling}. Prior methods for RS removal can be categorized as multi-frame and single-frame based. The former relies on temporal motion in consecutive frames for motion compensation, while the latter solely depends on a single frame for restoration. Single-frame RS correction is particularly challenging yet crucial in data-scarce situations. Many existing methods, including non-learning-based approaches~\cite{rengarajan2016bows,rengarajan2017unrolling,purkait2017rolling,zhuang2019learning,kandula2020deep}, rely on salient structures, such as straight lines. However, they may falter in cases where such salient structures are absent.

The majority of deep RS correction methods are multi-frame based~\cite{liang2008analysis,ringaby2012efficient,zhuang2017rolling,vasu2018occlusion}, with only a few considering a single frame as input. Among single-frame methods, Rengarajan~\emph{et al.}~\cite{rengarajan2017unrolling} directly learn the mapping between global shutter (GS) and RS frames, facing the challenge of large solution space for row-wise motion estimation. Zhuang~\emph{et al.}~\cite{zhuang2019learning} addressed this by estimating an additional depth map, although depth estimation from RS frames is inherently ill-posed. Recently, Yan~\emph{et al.}~\cite{yan2023deep} introduced a deep homography mixture model, achieving the current best performance by embedding motion in a subspace and learning coefficients to combine pre-learned motion flow bases.

{Recent advancements in generative models~\cite{brown2020language,goodfellow2020generative,ouyang2022training}, including Diffusion Models (DMs)~\cite{ho2020denoising,song2020score}, have significantly advanced the field of Artificial Intelligence. Notably, DMs excel in performing various motion-related tasks, notably in human motion generation~\cite{tevet2022human}, in estimating depth/optical flow~\cite{saxena2023surprising}, and in homography data rendering~\cite{li2024dmhomo}. Meanwhile, DMs have shown strength in solving ill-posed problems, such as image restoration~\cite{gao2023implicit,yinhuai2022zero}, where inferring accurate solutions from ambiguous data is difficult. Therefore, drawing on these insights and inspirations, in this work, we introduce a novel single-frame RS correction method based on DMs, namely \textbf{RS-Diffusion}.} DMs are adept at generating data from Gaussian distributions through multiple sampling steps and can be conditioned on additional information, such as images. Leveraging the transformative capabilities, we are capable of correcting RS images to GS images. Our framework, built on CFG~\cite{ho2022classifier} and DDIM~\cite{song2020denoising}, uses a downsampled RS image $I_{RS}$ as a conditioning element, concatenated with Gaussian noise. This input is processed by DMs in very few steps to produce a motion field $G$, which is then applied to remap $I_{RS}$ to eliminate RS artifacts. Additionally, we introduce a patch-attention module based on prior RS motion patterns to enhance results. We visualize our corrected result on the right side of the second row in Fig.~\ref{fig:teaser}.

On the other hand, high-quality datasets play a crucial role. To meet the demand, we adhere to two essential criteria~\cite{han2022realflow}: the \emph{label criterion}, which requires precise alignment of the RS correction motion as ground-truth labels between RS-GS pairs, and the \emph{realism criterion}, which ensures that both the image content and RS motion remain realistic. However, existing RS datasets often fall short of meeting these criteria. The synthesized dataset can provide accurate labels but violates realism criterion. In contrast, the captured images can suffice the realism, but the label is not accurate.

To resolve these issues, we introduce Intra Gyro Field (IGF) pipeline, leveraging Inertial Measurement Unit (IMU) gyroscope sensors to record camera rotations during capture. Specifically, we firstly achieve accurate frame-gyro data synchronization. Next, these rotations are translated into a series of homography matrices, further converted into motion fields as RS correction ground-truth (GT) labels. With the help of IGF, we create the \textbf{RS-Real} dataset, which satisfies both the label and the realism criteria in RS correction. The dataset contains 40,000 train and 1,000 test samples. A case is shown in Fig.~\ref{fig:teaser}, which shows a captured RS image, the IGF between the RS-GS pair and its GS image.


In summary, our diffusion model-based framework advances the state-of-the-art in single-image RS correction, while it could run inference in real-time speed, i.e., up to 28.1 ms per frame on one NVIDIA 2080Ti. The RS-Real dataset, containing high-quality training pairs, addresses the scarcity of qualified datasets in the RS task. Our contributions include:
\begin{itemize}
    
    \item We propose the first diffusion-based framework for single image rolling shutter removal, namely \textbf{RS-Diffusion}.
   
    \item We introduce a pipeline for correcting RS images with recorded IMU gyro readings, which delivers accurate rectified RS images as ground-truth labels for training and testing, yielding a realistic RS dataset, \textbf{RS-Real}.  
    
    \item Experiments show that our approach achieves state-of-the-art performance on public benchmarks, exhibiting both generalizability and applicability.
\end{itemize}

\section{Related Works}
\label{sec:related_works}

\subsection{Rolling Shutter Correction Methods}
Multi-frame rectification methods use spatial-temporal information, with investigations into per-pixel motion vectors~\cite{liang2008analysis}, image feature tracking~\cite{ringaby2012efficient}, and affine models~\cite{baker2010removing}, now enhanced by 3D data~\cite{zhuang2017rolling}. Deep learning advances include pixel-wise velocity estimation~\cite{liu2020deep} and optical flow for deep feature warping~\cite{fan2021sunet}.
Classical single-image methods mostly utilize salient lines for RS correction~\cite{rengarajan2016bows}. Deep learning for single RS image correction is less common but includes neural networks trained on paired images~\cite{rengarajan2017unrolling}, employing motion models and depth map estimations~\cite{zhuang2019learning}. A novel deep homography mixture model shows state-of-the-art results~\cite{yan2023deep}. We introduce diffusion models (DMs) for single RS rectification, demonstrating their effectiveness and generalizability in motion extraction.

\subsection{Diffusion Models} 
Diffusion models (DMs), based on stochastic diffusion processes~\cite{sohl2015deep}, efficiently transform data distributions through Gaussian transitions and iterative denoising using the data distribution gradient~\cite{song2019generative}. DDPM~\cite{ho2020denoising} employs discrete steps in this process, and Song~\emph{et al.}~\cite{song2020score} further refines the methodology via stochastic differential equations (SDE). DDIM~\cite{song2020denoising} accelerates reverse sampling through subsequence sampling and ordinary differential equations (ODEs). Conditioned data generation progresses with classifier-based~\cite{dhariwal2021diffusion,liu2023more} and classifier-free (CFG) techniques~\cite{ho2022classifier}. DMs also contribute to motion-centric tasks, from video generation~\cite{ni2023conditional}, video frame interpolation~\cite{danier2024ldmvfi}, object tracking~\cite{luo2024diffusiontrack} to Molecule generation~\cite{huang2023mdm}. Our pioneering work extends DMs to the domain of rolling shutter removal.

\begin{figure*}[t]
\begin{center}
    \includegraphics[width=0.9\linewidth]{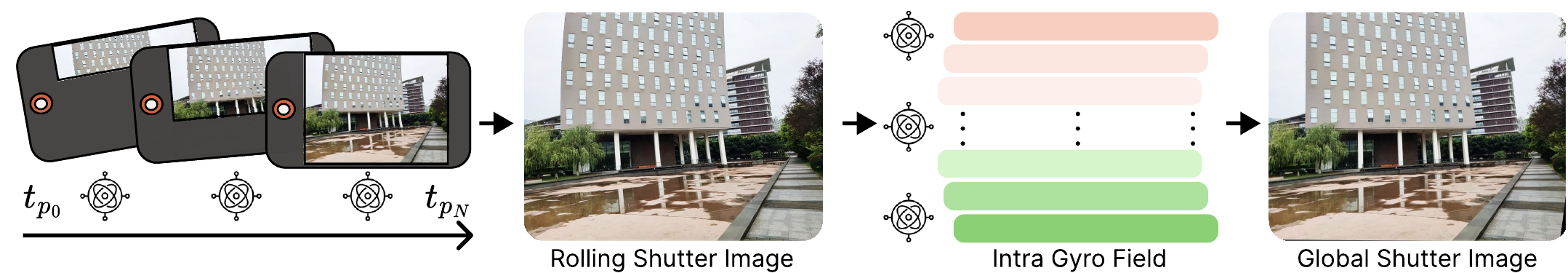}
\end{center}

\caption{ Rolling shutter image $\mathbf{I}_{RS}$, is introduced by high-frequency shake with a row-wise exposure CMOS camera. The gyroscope can accurately record these motions, which are then transformed into a motion field, $\mathbf{G} \in \mathbf{R}^{2 \times H \times W}$. This field is referred to as the Intra Gyro Field (IGF). With $\mathbf{G}$, we are able to correct $\mathbf{I}_{RS}$, resulting in a Global Shutter Image, $\mathbf{I}_{GS}$.}
\label{fig:intra_gyro_field}

\end{figure*}

\subsection{Gyroscope-based Motion Methods}

Gyroscopes are pivotal for diverse applications such as video stabilization~\cite{karpenko2011digital,bell2014non},  optical image stabilization (OIS)~\cite{liu2021deepois}, estimating homography/optical flow~\cite{li2021gyroflow,li2023gyroflow+}, and simultaneous localization and mapping (SLAM)~\cite{huang2018online}, with growing using in RS correction if gyroscope data is accessible during the capturing~\cite{mo2022imu}. In this work, instead of directly using online captured gyroscope data to promote the performances of different applications. We use the gyroscope data to create a dataset.   Our work utilizes gyroscope synchronization at the Android HAL for precise calibration~\cite{jia2013online,bloesch2014fusion}, enabling the creation of a \textbf{RS-Real} dataset for rolling shutter  research, incorporating authentic RS pairings for enhanced training and validation. Note that, our method doesn't need gyro readings during the model inference.


\section{Method}
\label{sec:method}

\subsection{Overview}
\label{sec:overview}

We present a pipeline named Intra Gyro Field (IGF) designed to create a realistic and precisely annotated dataset through combined engineering and algorithmic efforts, as discussed in Section \textbf{Intra Gyro Field}. Furthermore, we propose a novel and general framework based on diffusion models, incorporating a specialized patch attention (PA) module, which is aimed at correcting rolling shutter (RS) images to global shutter (GS) quality, as described in Section \textbf{Diffusion Models}.


\subsection{Intra Gyro Field}
\label{sec:intra-gyro-field}

Gyroscope records the relative camera 3D rotation in a relative time and it is possible to convert the gyroscope readings into a 2D motion field to achieve alignment between two consecutive frames~\cite{li2023gyroflow+}. In this chapter, we demonstrate how to leverage the gyroscope to correct RS effect for a single image and thus propose an Intra Gyro Field (IGF). Specifically, given the 3-axis angular velocity readings (row $\mathbf{v_{r}}$, pitch $\mathbf{v_{p}}$ and yaw $\mathbf{v_{y}}$) and relative time intervals $\Delta t$ between gyro readings. The 3-axis angular angles ($\mathbf{\angle r, \angle p, \angle y}$) can be computed as:
\begin{equation}
    \mathbf{\angle r} =  \mathbf{v_{r}} \cdot \Delta t, \quad \mathbf{\angle p} =  \mathbf{v_{p}} \cdot \Delta t, \quad \mathbf{\angle y} =  \mathbf{v_{y}} \cdot \Delta t.
\end{equation}
Then we utilize the Rodrigues Formula to convert rotation angles into the rotation matrix $\mathbf{R}(\Delta t)\in SO(3)$. In order to bridge the relationship between 3D camera pose and 2D image motion, we choose to use the homography matrix $\mathbf{H}$. The homography matrix represents the relationship between two different perspectives of the same scene, more specifically, the conditions for its validity are satisfied when the camera motion is purely rotational or when the contents lie in the same plane~\cite{hartley2003multiple}. Theoretically, given 3D rotation matrix $\mathbf{R}(\Delta t)$ and translation vector $\mathbf{t}(\Delta t)$), we can represent $\mathbf{H}$ as:

\begin{equation}
     {\mathbf{H}=  \mathbf{K}\left(\mathbf{R}(\Delta t)+\mathbf{t}(\Delta t) {\frac{\mathbf{n}^T}{d}}\right) \mathbf{K}^{-{1}}},
     \label{eq:camera_pose_to_homo}
\end{equation}
where $\mathbf{{n}}^T$ represents the normal vector of a certain plane, $d$ is the distance between the camera center and the plane, while $\mathbf{K}$ is the camera intrinsic matrix. 

\begin{figure*}[t]
\begin{center}
    \includegraphics[width=1\linewidth]{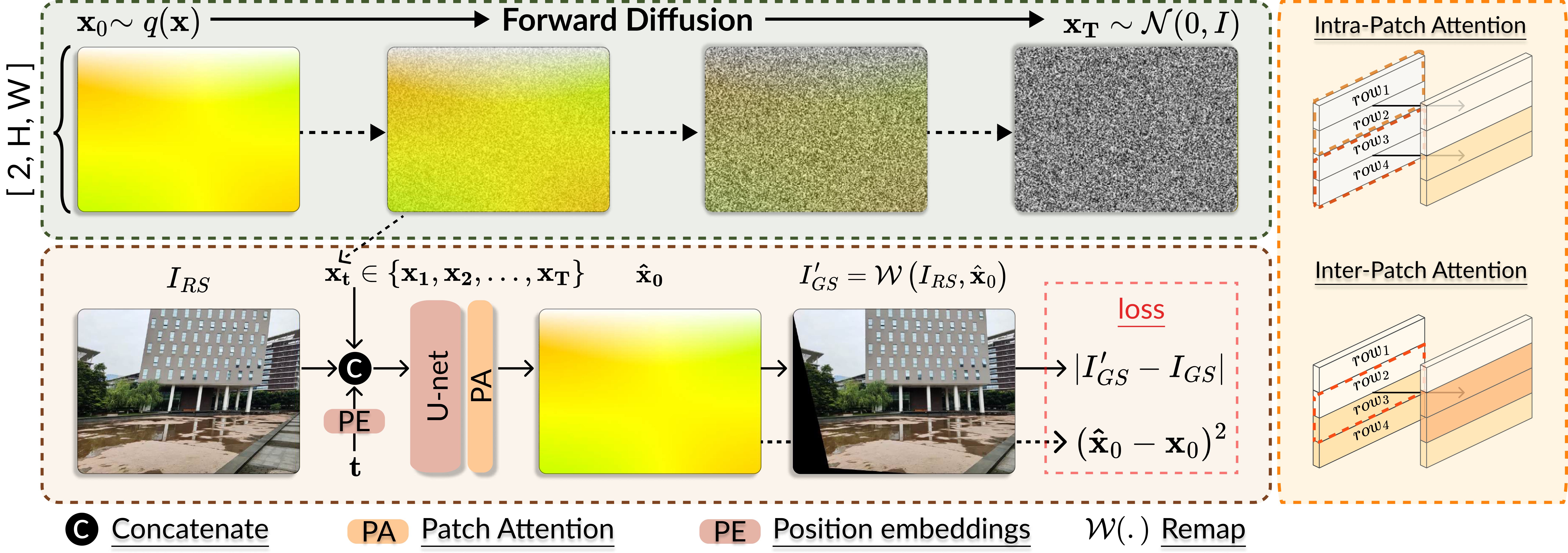}
\end{center}

\caption{Illustration of the framework: During training, $\mathbf{x}_0$ undergoes forward diffusion to become $\mathbf{x}_t$. The network $\mathbf{\theta}$ processes the concatenated input. The Patch-Attention module, which includes both Intra-Patch and Inter-Patch attention mechanisms, enhances the relationships between patches. The resulting output, $\mathbf{\hat{x}_0}$, can be used to correct $\mathbf{I}_{RS}$. The loss function comprises the MSELoss, calculated between $\mathbf{\hat{x}_0}$ and $\mathbf{x}_0$, and the photometric loss, computed between the GT GS image and $\mathbf{I^{\prime}}_{GS}$.}
\label{fig:rs_diffusion_ppl}

\end{figure*}

\begin{figure}[t]
\begin{center}
    \includegraphics[width=1\linewidth]{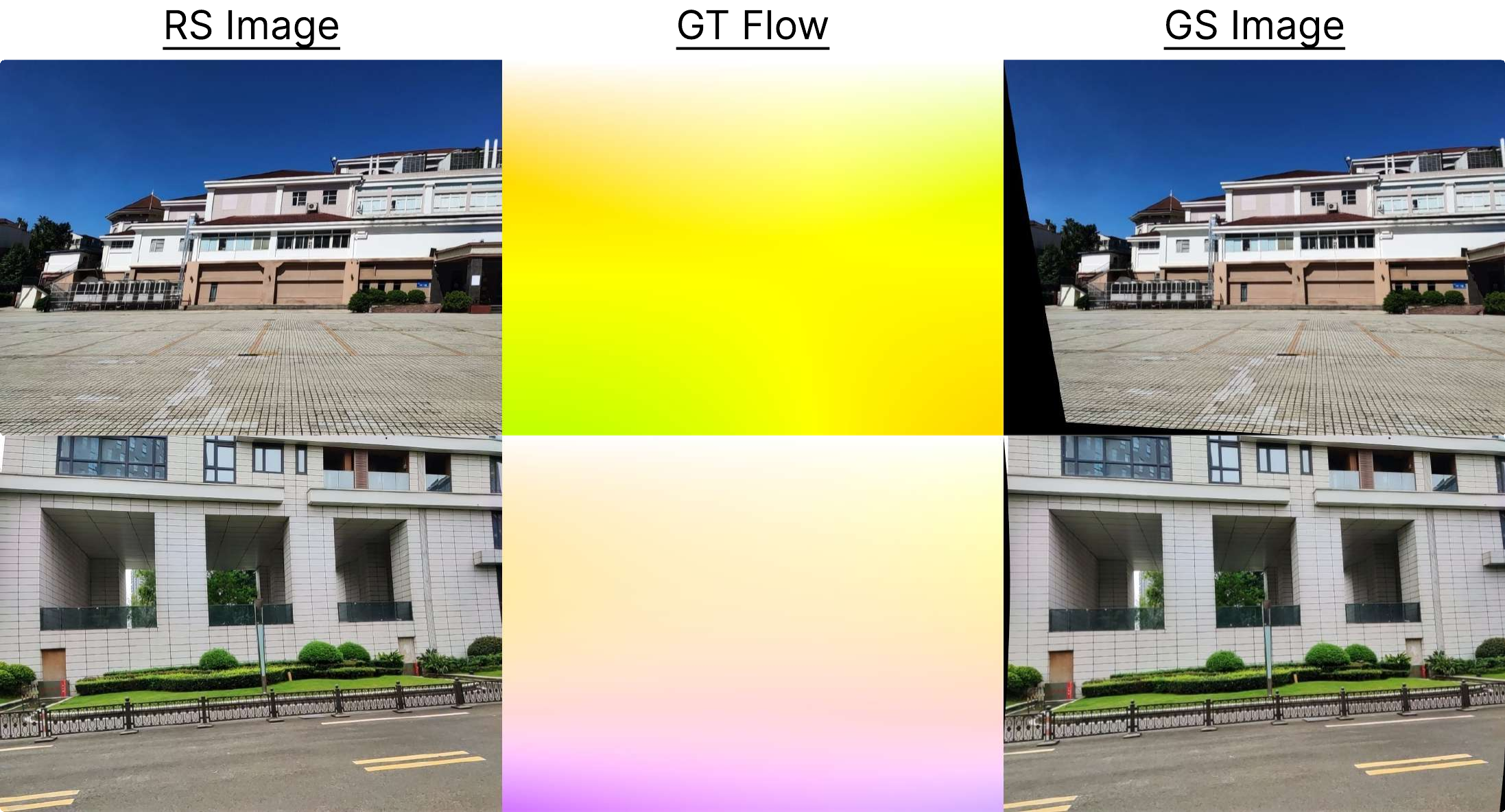}
\end{center}

\caption{A glance at the \textbf{RS-Real} dataset reveals that it contains data pairs featuring various RS motion patterns and different intensities, all of which are captured in diverse scenes.}
\label{fig:gyro_rs_dataset}
\end{figure}

In this work, we do not incorporate translations and only use gyroscope data to model the homography matrix, the reasons are threefold. Firstly, the rolling shutter effect caused by camera shake occurs primarily due to rotational movements~\cite{karpenko2011digital}. Secondly, even though translations can be gathered from accelerometer data, they are significantly less accurate than rotational measurements~\cite{forssen2010rectifying,joshi2010image}. Lastly, according to  Eq.~\ref{eq:camera_pose_to_homo}, we can find that translation is correlated to depth, but accurately estimating depth maps can be another non-trivial problem. As a result, we can formulate a rotational-only homography:

\begin{equation}
     \mathbf{H}=  \mathbf{K}\left(\mathbf{R}(\Delta t)\right) \mathbf{K}^{-{1}}.
     \label{eq:rotation_to_homo}
\end{equation}

The gyroscope records $N$ readings during the capture of one image, it thus can produce $N$ camera rotation matrices ($\mathbf{R}_{1}(\Delta t), \mathbf{R}_{2}(\Delta t), \cdots, \mathbf{R}_{N}(\Delta t)$). If we split the image from top to bottom into $N$ patches ($\mathbf{p}_1, \mathbf{p}_2, \cdots, \mathbf{p}_N$), these rotation matrices correspond to the motion between the first row and bottom row of patches as illustrated in Fig.~\ref{fig:intra_gyro_field}. Moreover, the inter-patch motion is approximately to be smooth~\cite{liu2021deepois,dai2016rolling}, so we can apply the spherical linear interpolation (SLERP) to interpolate the motion to avoid the discontinuities across row patches. Subsequently, we follow Eq.~\ref{eq:rotation_to_homo} to produce $N$ homography matrices ($\mathbf{H}_1, \mathbf{H}_2, \cdots, \mathbf{H}_N$) to model the 2D relationship between patches. To this end, we follow previous method~\cite{li2021gyroflow} to convert the array of homography matrix into a motion field, it is achieved by transforming grid points by their corresponding homography and subtracting them. Lastly, this motion field is IGF, denoted as $\mathbf{G} \in \mathbf{R}^{2 \times H \times W}$, which is capable of correcting the RS effect within an image:

\begin{equation}
     \mathbf{I}_{GS}=  \mathcal Remap (\mathbf{I}_{RS}, \mathbf{G}).
     \label{eq:rs-to-gs}
\end{equation}

\noindent\textbf{Dataset Collecting.} Our proposed IGF effectively addresses the challenges found in previous datasets, such as unrealistic data distributions and spatio-temporal synchronization issues. By recording the RS effect with a gyroscope and subsequently recovering from it, our approach circumvents synchronization challenges by using only one camera at a time. As a result, we can flexibly capture a diverse range of realistic scenes, spanning various indoor and outdoor environments, and subjecting the camera to multiple motion patterns and speeds. This method enables us to gather a comprehensive dataset of realistic RS images, denoted as $\mathbf{X}_{RS}=\left\{\mathbf{I}_{RS}^{0}, \mathbf{I}_{RS}^{1},\ldots,\mathbf{I}_{RS}^{k}\right\}$, along with their corresponding IGFs, $\mathbf{X}_{IGF}=\left\{\mathbf{G}^{0}, \mathbf{G}^{1},\ldots,\mathbf{G}^{k}\right\}$. After applying RS removal using Eq.~\ref{eq:rs-to-gs}, we obtain the set of GS images, $\mathbf{X}_{GS}=\left\{\mathbf{I}_{GS}^{0}, \mathbf{I}_{GS}^{1},\ldots,\mathbf{I}_{GS}^{k}\right\}$. We refer to this  dataset as \textbf{RS-Real}. Examples from the dataset are illustrated in Fig.~\ref{fig:gyro_rs_dataset}.

\subsection{Diffusion Models}
\label{sec:diffusion_models}

To correct distortions resulting from the rolling shutter effect in general cases, we utilize diffusion models adept at managing the distributional transformations of images captured under both RS and GS conditions. 

Specifically, we leverage the diffusion models as proposed by Sohl-Dickstein~\emph{et al.}~\cite{sohl2015deep} and Ho~\emph{et al.}~\cite{ho2020denoising}, which define a Markov chain process over $T$ steps to incrementally introduce Gaussian noise into the original data distribution $\mathbf{x}_0 \sim q(\mathbf{x})$. This stepwise infusion of noise generates a sequence of increasingly distorted states ${\mathbf{x}_1, \ldots, \mathbf{x}_T}$, defined as forward diffusion, and is mathematically represented as:
\begin{equation}
q\left(\mathbf{x}_{1: T} \mid \mathbf{x}_0\right)=\prod_{t=1}^T q\left(\mathbf{x}_t \mid \mathbf{x}_{t-1}\right).
\end{equation}
Following this, a denoising model, denoted as $\mathbf{\theta}$, is then trained to reverse the diffusion process $p_\theta\left(\mathbf{x}_{t-1} \mid \mathbf{x}_t\right)$ to construct desired data samples from isotropic Gaussian noise $\mathbf{x}_T \sim \mathcal{N}(\mathbf{0}, \mathbf{I})$:
\begin{equation}
p_\theta\left(\mathbf{x}_{0: T}\right)=p\left(\mathbf{x}_T\right) \prod_{t=1}^T p_\theta\left(\mathbf{x}_{t-1} \mid \mathbf{x}_t\right).
\end{equation}
This inversion transforms the signal from the noise-dominant Gaussian distribution back into the target data distribution. To refine this reconstruction process and improve the quality of the generated samples, we can integrate conditional variables $\mathbf{y}$ into the model as described by CFG~\cite{ho2022classifier}. These conditions are merged with the noisy data transitions in the model:
\begin{equation}
p_\theta\left(\mathbf{x}_{t-1} \mid \mathbf{x}_t, \mathbf{y}\right)=\mathcal{N}\left(\mathbf{x}_{t-1} ; \boldsymbol{\mu}_\theta\left(\mathbf{x}_t, t, \mathbf{y}\right), \sigma_t^2 \mathbf{I}\right).
\end{equation}
By incorporating supplementary conditions, we are able to exert enhanced control over the sample generation phase, which results in an improvement in the fidelity of the final reconstructions.





\subsection{Rolling Shutter Removal Module}
\label{sec:rs-correct-module}
Our framework is built upon CFG~\cite{ho2022classifier} and DDIM~\cite{song2020denoising}. We illustrate the pipeline in the Fig.~\ref{fig:rs_diffusion_ppl}, we designate IGFs as the initial data distribution, $\mathbf{x}_0$, and the RS images $\mathbf{I}_{RS}$ as conditions $\mathbf{y}$. This configuration endows the model with an ``\textit{image-to-motion}" capability. During the training process, $\mathbf{x}_0$ is noised via forward diffusion to $\mathbf{x}_t$, then we concatenate the $\mathbf{I}_{RS}$, $\mathbf{x}_t$ and time embedding to fed into the U-net network $\theta$. It is imperative to mention that this concatenation is introduced at various layers within the network. Finally, the output of $\theta$ is processed through the Patch-Attention module, resulting in the generated outputs $\hat{\mathbf{x}}_0$. 

\noindent\textbf{Patch-Attention Module.} As described in previous sections, we observe that the motion patterns in RS images are consistently correlated between rows. In practice, rows are often regrouped into patches to mitigate the effects of RS. Capitalizing on this a priori, we propose an attention block to foster the inter-relationships between these patches. Our attention block operates in two stages: Firstly, the intra-patch attention phase involves evenly splitting the input into non-overlapping patches. As exemplified on the right side of Fig.~\ref{fig:rs_diffusion_ppl}, the feature is divided into patches (4 rows into 2 patches in this example). Within each patch, self-attention is employed to process features internally. Subsequently, the inter-patch attention stage is applied. In this phase, self-attention mechanisms are utilized to facilitate the exchange of information between consecutive rows across different patches. This approach ensures that the relationship between {adjacent} patches is effectively enhanced.


\noindent\textbf{Loss Function.} After computing $\hat{\mathbf{x}}_0$, we first apply the mean squared error (MSE Loss) as follows:
\begin{equation}
    \ell_{mse} = (\mathbf{\hat{x}}_0 - \mathbf{x}_0)^2.
\end{equation}
In addition to the standard MSE Loss, which facilitates learning the distribution transformation from RS images to their respective IGF, we further propose to constrain the network with an extra conditional loss. Specifically, we warp the original RS images $\mathbf{I}_{RS}$ via the computed IGF $\mathbf{\hat{x}}_0$ to produce corrected RS images $\mathbf{I}_{G S}^{\prime}$:
\begin{equation}
    \mathbf{I}_{G S}^{\prime}=\mathcal{W}\left(\mathbf{I}_{R S}, \hat{\mathbf{x}}_0\right),
\end{equation}
where $\mathcal{W}(.)$ represents the remapping operation. Then we calculate the photometric loss between $\mathbf{I}_{G S}^{\prime}$ and $\mathbf{I}_{GS}$:
\begin{equation}
    \ell_{pl} = |\mathbf{I}_{G S}^{\prime} - \mathbf{I}_{GS}|.
\end{equation}
Consequently, the overall loss can be computed as a dynamically weighted sum. In other words, it continually adjusts $\ell_{pl}$ to be equal to $\ell_{mse}$, formulated as:
\begin{equation}
    \ell_{overall} = \ell_{mse} + \frac{\left|\ell_{mse}\right|}{\left| \ell_{pl} \right|} \cdot \ell_{pl}.
\end{equation}

\section{Experiments}
\label{sec:experiments}
We begin by outlining existing and our proposed datasets in Section \textbf{Dataset}. Our method's performance is compared against others on varied benchmarks in Section \textbf{Quantitative Comparison}, with qualitative evaluations presented in Section \textbf{Qualitative Comparison}. An in-depth analysis of our framework's intricate designs is described in Section \textbf{Ablation Studies}. Finally, we list implementation details and demonstrate the utility of our framework in a video stabilization application~\cite{liu2013bundled} within the \textbf{supplementary materials}.

\subsection{Dataset}
\label{sec:datasets}
In our experiments, we validate the proposed RS correction method on 3 datasets, including one long-standing public datasets and two new ones for comprehensive evaluation. To ascertain the reliability of our approach, we synchronize our evaluations with established RS image correction techniques~\cite{rengarajan2017unrolling, kandula2020deep,yan2023deep}, utilizing the \textbf{Building} dataset~\cite{xiao2010sun, 2003ETHbuilding, 2007oxford_building} as a standard benchmark. Additionally, we extend our test scope with a challenging dataset \textbf{RS-Homo} from Yan~\emph{et al.}~\cite{yan2023deep}, crafted to simulate adverse conditions like low lighting and scant textures, critical for testing diffusion model robustness. A notable contribution of this paper is the \textbf{RS-Real} dataset, detailed in Section~\ref{sec:intra-gyro-field}, which exploits sensor information during image capture to address common synchronization challenges and offers a new methodology for data gathering in RS scenarios.

\subsection{Quantitative Comparison}
\label{sec:quanti_comp}
In our study, we evaluate our RS image correction method using key metrics for visual quality and motion accuracy. Visual comparisons are done via the Peak Signal-to-Noise Ratio (PSNR) and the Structural Similarity Index Measure (SSIM) against GT GS images. The motion-based assessment considers the Endpoint Error (EPE) for correction flow accuracy. Pixels on black edges, lacking information, are omitted from our analysis.

\begin{figure*}[t]
\begin{center}
    \includegraphics[width=1.0\linewidth]{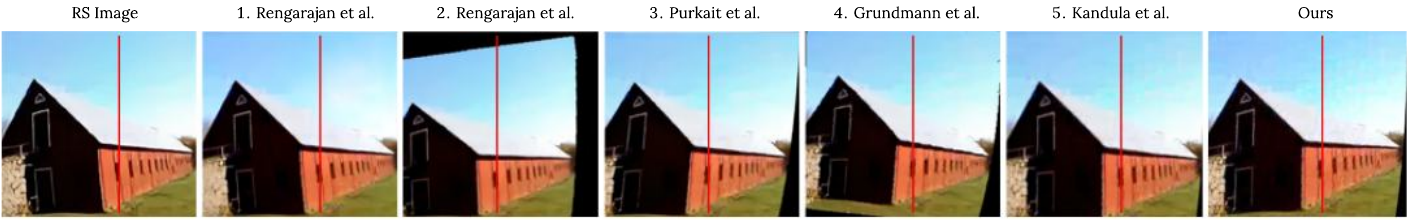}
\end{center}

\caption{Comparison with existing methods: 1. Rengarajan~\emph{et al.}~\cite{rengarajan2016bows}, 2. Rengarajan~\emph{et al.}~\cite{rengarajan2017unrolling},
3. Purkait~\emph{et al.}~\cite{purkait2017rolling}, 4. Grundmann~\emph{et al.}~\cite{grundmann2012calibration} and 5. Kandula~\emph{et al.}~\cite{kandula2020deep} on a RS building image. Red vertical lines to highlight correction results.}
\label{fig:crop-comparison-qualitative}

\end{figure*}

\begin{table}[h]
\centering
\def\temptablewidth{0.45\textwidth}
{\rule{\temptablewidth}{1pt}}  

\begin{tabular*}{\temptablewidth}{@{\extracolsep{\fill}}l|ccc}
\hline
Method&PSNR(dB)$\uparrow$ &SSIM$\uparrow$& EPE$\downarrow$\\ 
\hline
1. Rengarajan~\emph{et al.}&29.82 &0.67&11.89\\
2. Purkait~\emph{et al.}&29.22&0.55&8.32\\


3. Grundmann~\emph{et al.} &32.57 &0.72 &3.34\\

4. Rengarajan~\emph{et al.}&32.25&0.70&3.76\\

5. Kandula~\emph{et al.}&32.85&0.73&2.84\\
6. Yan~\emph{et al.}& {33.34} & {0.75} & {1.25}\\
\hline
RS-Diffusion& \textbf{34.92} & \textbf{0.79} & \textbf{0.90}\\
\hline
\end{tabular*}

{\rule{\temptablewidth}{1pt}}

\caption{Comparison of PSNR, SSIM, and EPE between our method and existing ones: 1. Rengarajan~\emph{et al.}~\cite{rengarajan2016bows}, 2. Purkait~\emph{et al.}~\cite{purkait2017rolling}, 3. Grundmann~\emph{et al.}~\cite{grundmann2012calibration}, 4. Rengarajan~\emph{et al.}~\cite{rengarajan2017unrolling}, 5. Kandula~\emph{et al.}~\cite{kandula2020deep} and 6. Yan~\emph{et al.}~\cite{yan2023deep} on \textbf{Building} dataset.}

\label{table:compare with pre} 

\end{table}

\noindent\textbf{Building dataset}. Our study compares with traditional and deep learning-based RS correction methods, as shown in Table~\ref{table:compare with pre}. Approaches using curve detection~\cite{rengarajan2016bows, purkait2017rolling} to discern motion patterns struggle in low-structure environments, showing subpar PSNR and SSIM, alongside elevated EPE. The homography mixtures method~\cite{grundmann2012calibration} fares better with its video sequence foundation and refined feature correspondences. However, learning methods~\cite{rengarajan2017unrolling, kandula2020deep} have proven robust to strong outliers and complex camera motions. Yan~\emph{et al.} model~\cite{yan2023deep}, merging homography mixtures with learning, showcases significant strides in metric performance due to constrained motion spaces from learned bases. Our proposed technique outperforms existing methods, setting a new state-of-the-art (SOTA) across evaluated metrics and emphasizing its effectiveness.


\begin{table}[h]
\centering
\def\temptablewidth{0.42\textwidth}
{\rule{\temptablewidth}{1pt}}  

\begin{tabular*}{\temptablewidth}{@{\extracolsep{\fill}}c|ccc}
\hline
Method&PSNR(dB)$\uparrow$ &SSIM$\uparrow$& EPE$\downarrow$\\ 
\hline
Yan~\emph{et al.}& {26.15} & {0.77} & {4.10}\\
\hline
RS-Diffusion& \textbf{36.60} & \textbf{0.94} & \textbf{1.02}\\
\hline
\end{tabular*}

{\rule{\temptablewidth}{1pt}}
\caption{Comparison of PSNR, SSIM, and EPE between our method and Yan~\emph{et al.}~\cite{yan2023deep} on \textbf{RS-Homo}.}

\label{table:deep-hx-synth-data} 

\end{table}

\noindent\textbf{RS-Homo dataset.} The results in Table~\ref{table:deep-hx-synth-data} show our diffusion models-based framework substantially improves upon Yan \emph{et al.}'s state-of-the-art single-image model. Demonstrating great proficiency in managing challenging scenarios, our proposed method surpasses specifically designed architectures, enhancing the capability of diffusion models to address rolling shutter (RS) effects effectively.

\begin{table}[h]
\centering
\def\temptablewidth{0.42\textwidth}
{\rule{\temptablewidth}{1pt}}  

\begin{tabular*}{\temptablewidth}{@{\extracolsep{\fill}}c|ccc}
\hline
Method&PSNR(dB)$\uparrow$ &SSIM$\uparrow$& EPE$\downarrow$\\ 
\hline
Yan~\emph{et al.}& {18.48} & {0.55} & {4.18}\\
\hline
RS-Diffusion & \textbf{22.02} & \textbf{0.69} & \textbf{2.12}\\
\hline
\end{tabular*}

{\rule{\temptablewidth}{1pt}}
\caption{Comparison of PSNR, SSIM, and EPE between our method and Yan~\emph{et al.}~\cite{yan2023deep} on \textbf{RS-Real}.}

\label{table:intra-gyro-field-data} 

\end{table}

\noindent\textbf{RS-Real dataset}. Our dataset, designed for realism in content, RS-motion, and label accuracy, comprises 40,000 training and 1,000 test pairs across diverse scenes. It includes RS and GS images along with GT flow. We benchmark against Yan~\emph{et al.}~\cite{yan2023deep} by retraining their model using the default settings on our trainset. The results confirm that our approach consistently outperforms theirs in both photometric and motion measures as demonstrated in Table~\ref{table:intra-gyro-field-data}.

\begin{figure*}[h]
\begin{center}
    \includegraphics[width=1\linewidth]{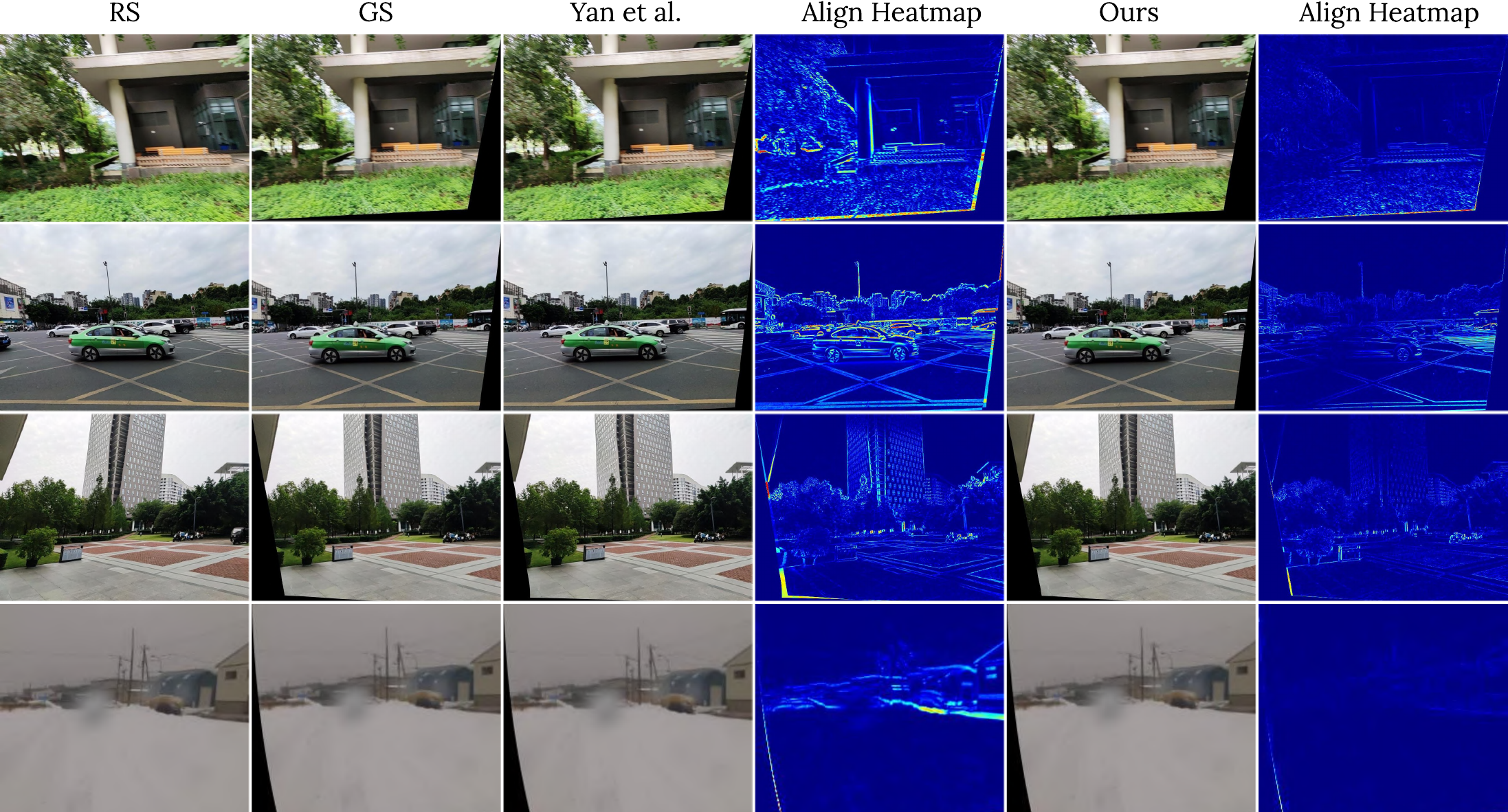}
\end{center}

\caption{Comparison on \textbf{RS-Real} and \textbf{RS-Homo} dataset. Column 1 shows the input RS image and Column 2 shows the ground-truth GS image. Column 3 and 5 are results by Yan~\emph{et al.}~\cite{yan2023deep} and our method with the alignment heatmaps that use darker shades to indicate greater similarity compared to ground-truth GS images.}
\label{fig:rs-real-qualitative}

\end{figure*}

\subsection{Qualitative Comparison}
\label{sec:quali_comp}

Our method is evaluated against current methods on different benchmarks. We follow the settings used by Yan~\emph{et al.}~\cite{yan2023deep} to compare with existing methods~\cite{rengarajan2017unrolling,rengarajan2016bows,purkait2017rolling,grundmann2012calibration,kandula2020deep} in Fig.~\ref{fig:crop-comparison-qualitative}, and ours outperforms the others. In addition, due to limited open-source options, we compare with Yan~\emph{et al.}~\cite{yan2023deep} method on the remaining datasets. We visualize RS and ground-truth GS images in the first two columns, then present Yan~\emph{et al.} results paired with alignment heatmaps that use darker shades to indicate greater similarity compared to GT images. We then display our own results alongside their corresponding alignment heatmaps, illustrating our approach's consistency with benchmarks and precision in alignment. This visual comparison highlights the effectiveness of our methodology.


\noindent\textbf{RS-Real and RS-Homo Datasets}.
In the first three rows of  Fig.~\ref{fig:rs-real-qualitative}, we present the results on the \textbf{RS-Real} dataset, while the last row shows the results on the \textbf{RS-Homo} dataset. Our approach consistently outperforms the method of Yan~\emph{et al.}~\cite{yan2023deep} across varied RS challenges, including right-skewed images, dynamic scenes with moving cars, complex motion patterns from quick device movement, and low-texture scenarios, proving its robustness and superior correction capabilities.

\subsection{Ablation Studies}
\label{sec:abla_study}

We evaluate our framework design through experiments, starting with comparisons under diverse generative objectives, paired with analytical analyses. The effectiveness of the Patch-Attention Block is then scrutinized.


\subsubsection{Predicting IGF vs. GS Image}

\begin{table}[H]
\centering
\def\temptablewidth{0.37\textwidth}
{\rule{\temptablewidth}{1pt}}  

\begin{tabular*}{\temptablewidth}{@{\extracolsep{\fill}}c|cc}
\hline
Method &PSNR(dB)$\uparrow$ &SSIM$\uparrow$ \\ 
\hline
\textit{``image-to-image"} & 16.73 & 0.47\\
\hline
\textit{``image-to-motion"} & \textbf{20.78} & \textbf{0.63}\\
\hline
\end{tabular*}

{\rule{\temptablewidth}{1pt}}
\caption{Comparison between different generating objects. ``image-to-image'' indicates transforming RS images to GS. ``image-to-motion" represents predicting IGFs from RS.}

\label{table:backbone-design} 

\end{table}
 
We contrast our ``image-to-motion" pipeline with traditional ``image-to-image" methods that use diffusion models to convert RS images to GS images at a fixed resolution of 256$\times$256, later upscaled to 600$\times$800 for visual metric comparison with GT GS images, as shown in Table~\ref{table:backbone-design}. Our approach outperforms the conventional framework in the RS removal task for two main reasons: 1) It is not constrained by diffusion model resolutions, since we can upsample IGF to correct RS images at their original size; 2) While ``image-to-image" models learn the joint probability distribution between RS and GS images, our method gains additional improvements by learning from the distribution involving IGFs, RS and GS images, thus leveraging more information for enhanced results.

\subsubsection{Rolling-Shutter Patch Attention}
As demonstrated in Table~\ref{table:pa-atten}, our proposed Patch Attention module proves to be effective, as it not only aggregates features within individual patches but also facilitates interaction between features across different patches. The experimental results align well with our prior understanding of RS motion.

\begin{table}[h]
\centering
\def\temptablewidth{0.43\textwidth}
{\rule{\temptablewidth}{1pt}}  

\begin{tabular*}{\temptablewidth}{@{\extracolsep{\fill}}cc|ccc}
\hline
Intra &Inter &PSNR(dB)$\uparrow$ &SSIM$\uparrow$ &EPE$\downarrow$ \\ 
\hline
 & & 20.78 & 0.63 & 2.62\\
\hline
$\checkmark$ & & 21.10 & 0.65 & 2.54\\
\hline
$\checkmark$ & $\checkmark$ & \textbf{22.02} & \textbf{0.69} & \textbf{2.12}\\
\hline

\end{tabular*}

{\rule{\temptablewidth}{1pt}}
\caption{The effectiveness of Patch Attention. Intra refers to intra-patch attention and Inter indicates inter-patch attention.}

\label{table:pa-atten} 

\end{table}

\section{Conclusion}
\label{sec:conclusion}

In this work, we have presented RS-Diffusion, the first Diffusion Model based approach for single frame rolling shutter rectification. We have captured a novel dataset, namely RS-Real, to accomplish this task. The RS-Real dataset contains captured RS images, and the corresponding ground-truth GS images can be created according to synchronized gyroscope data that recorded during the RS frame capturing, yielding RS-GS image pairs. Uniquely, this dataset fulfills both accuracy and authenticity requirements for RS research. In addition, we have presented RS-Diffusion for real-time RS correction using just one RS frame. We have achieved state-of-the-art performances when compared to previous single RS correction methods. We publicly share our code and dataset with the community at https://github.com/lhaippp/RS-Diffusion. 

\section{Acknowledgements}
This work was supported in part by National Natural Science Foundation of China (NSFC) under Grant Nos. 62372091, 62071097 and in part by Sichuan Science and Technology Program under Grant Nos. 2023NSFSC0462, 2023NSFSC0458, 2023NSFSC1972.

\bibliography{aaai25}

\end{document}